\documentclass[twocolumn]{article}

\usepackage{graphicx}
\usepackage{amsmath}
\usepackage{amsfonts}
\usepackage{amssymb}
\usepackage{float}
\usepackage{hyperref}
\usepackage[top=1in, bottom=1in, left=1in, right=1in]{geometry}

\title{Multimodal Sentiment Analysis on CMU-MOSEI Dataset using Transformer-based Models}
\author{
    Jugal Gajjar\\George Washington University\\jugal.gajjar@gwu.edu
    \and
    Kaustik Ranaware\\George Washington University\\k.ranaware@gwu.edu
}
\date{}

\begin{document}

\maketitle

\begin{abstract}
    This project performs multimodal sentiment analysis using the CMU-MOSEI dataset, using transformer-based models with early fusion to integrate text, audio, and visual modalities. We employ BERT-based encoders for each modality, extracting embeddings that are concatenated before classification. The model achieves strong performance, with 97.87\% 7-class accuracy and a 0.9682 F1-score on the test set, demonstrating the effectiveness of early fusion in capturing cross-modal interactions. The training utilized Adam optimization (lr=1e-4), dropout (0.3), and early stopping to ensure generalization and robustness. Results highlight the superiority of transformer architectures in modeling multimodal sentiment, with a low MAE (0.1060) indicating precise sentiment intensity prediction. Future work may compare fusion strategies or enhance interpretability. This approach utilizes multimodal learning by effectively combining linguistic, acoustic, and visual cues for sentiment analysis.
\end{abstract}

\tableofcontents

\section{Introduction}
\label{sec_intro}

Traditionally, sentiment analysis, the task of analyzing emotions, was primarily conducted using textual data. However, we humans express our emotions not just with words, but through other non-verbal ways like facial expressions and tone of our voice as well \cite{russell2003facial}. These cues carry important contextual information that cannot be captured purely using text-based models \cite{poria2017review}. As a result, multimodal sentiment analysis, which uses more than one type of data to determine emotions has emerged as a more comprehensive approach to helping machines understand human emotions \cite{poria2017review}.

In this project, we perform multimodal sentiment analysis using the CMU-MOSEI dataset \cite{zadeh2018multimodal}, one of the largest and most diverse datasets available for this task. Using this dataset, our goal is to build a model that can accurately predict sentiments using information from all three modalities. We use transformer-based encoding, specifically BERT encodings across text, audio, and visual inputs. We apply an early fusion strategy combining the processed features from each data type at an embedding level before passing them to the prediction transformer.

Through this approach, we aim to demonstrate the effectiveness of early fusion and transformer architectures in decoding multimodal sentiment patterns without relying on modality-specific models.

\section{About CMU-MOSEI}
\label{sec_data}

For this project, we used the CMU-MOSEI (Multimodal Opinion Sentiment and Emotion Intensity) dataset \cite{zadeh2018multimodal}, a benchmark dataset created and maintained by Carnegie Mellon University. It is one of the largest benchmark dataset for multimodal sentiment analysis. It contains rich data aligned across three modalities—textual, visual, and acoustic.

The CMU-MOSEI dataset contains over 23,000 annotated sentence-level video segments from about 1,000 YouTube speakers \cite{zadeh2018multimodal}. These segments are chosen randomly from various topics and monologue videos \cite{zadeh2018multimodal}. All the data is transcribed and punctuated properly \cite{zadeh2018multimodal}. All data is annotated for both sentiment polarity and emotion intensity \cite{zadeh2018multimodal}, making it a great choice for multimodal learning.

Table 1 provides the dataset statistics \cite{zadeh2018multimodal}.

\begin{table}[h!]
    \centering
    \begin{tabular}{|p{0.7\linewidth}|l|}
        \hline
        \textbf{Feature} & \textbf{Name} \\
        \hline
        Total number of sentences & 23453 \\
        Total number of videos & 3228 \\
        Total number of distinct speakers & 1000 \\
        Total number of distinct topics & 250 \\
        Average number of sentences in a video & 7.3 \\
        Average length of sentences in seconds & 7.28 \\
        Total number of words in sentences & 447143 \\
        Total of unique words in sentences & 23026 \\
        Total number of words appearing at least 10 times & 3413 \\
        Total number of words appearing at least 20 times & 1971 \\
        Total number of words appearing at least 50 times & 888 \\
        \hline
    \end{tabular}
    \caption{CMU-MOSEI dataset statistics.}
    \label{tab:data_stat}
\end{table}

Figure 1 shows the diversity of topics of videos in CMU-MOSEI, displayed as a word cloud \cite{zadeh2018multimodal}. Larger words indicate more videos from that topic.

\begin{figure}[h!]
    \centering
    \includegraphics[width=\linewidth]{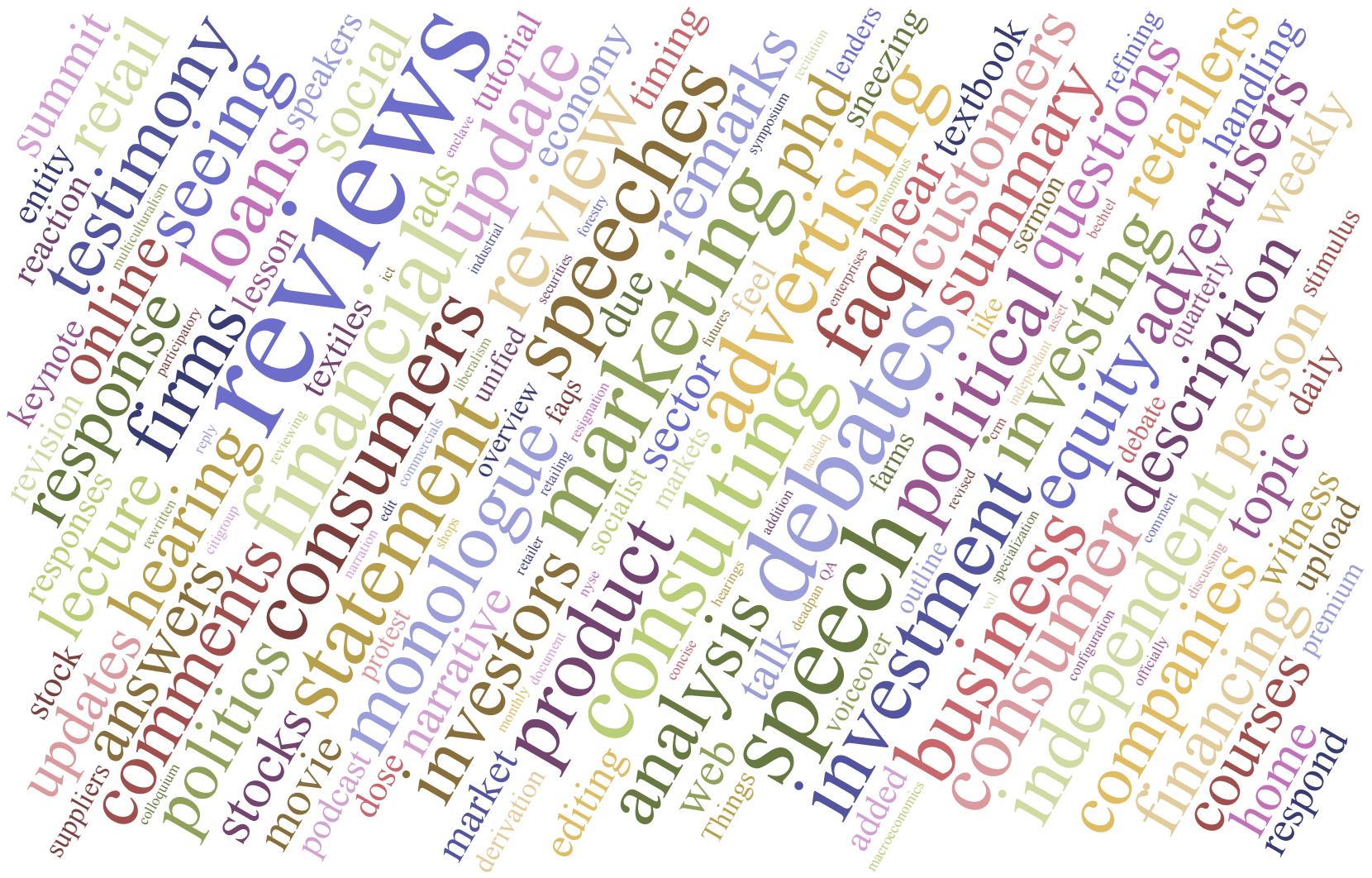}
    \caption{Word cloud visualizing topic diversity in CMU-MOSEI.}
    \label{fig:word_cloud}
\end{figure}

\subsection{Data Features}

The textual features in the dataset were extracted from the manual transcriptions using the Glove word embeddings \cite{pennington2014glove}. Both words and audio were aligned using the P2FA forced alignment model \cite{yuan2008speaker}. The visual and acoustic modalities are aligned using interpolation.

For the creation of visual features, the facial expressions were extracted from the video frames using MTCNN \cite{zhang2016joint} followed by the Facial Action Coding System (FACS) \cite{ekman1980facial}. Then, the static faces were classified into six basic emotions—anger, disgust, fear, happiness, sadness, and surprise using Emotient FACET \cite{imotions2017facial}. The data contains a set of 68 facial landmarks, 20 facial shape parameters, Histogram of Gradient (HoG) features, head pose, head orientation, and eye gaze extracted using MultiComp OpenFace \cite{baltrusaitis2016openface}. Finally, they extracted face embeddings using DeepFace \cite{taigman2014deepface}, FaceNet \cite{schroff2015facenet}, and SphereFace \cite{liu2017sphereface}.

The acoustic data was extracted using COVAREP software \cite{degottex2014covarep}. The extracted features include 12 Mel-frequency cepstral coefficients (MFCCs), pitch, voiced/unvoiced segmenting features \cite{drugman2011joint}, glottal source parameters \cite{drugman2012detection,alku2002normalized,alku1997parabolic}, peak slope parameters, and maxima dispersion quotients \cite{kane2013wavelet}, all of which are related to emotions and tone of speech.

The sentiment labels in the dataset range from -3 to +3 (-3, -2, -1, 0, +1, +2, +3), where -3 means strong negative sentiment, 0 means neutral, and +3 means strong positive sentiment \cite{zadeh2018multimodal}.

The dataset is already preprocessed and aligned across all three modalities. It is available for download through CMU Multimodal Data SDK at \url{https://github.com/CMU-MultiComp-Lab/CMU-MultimodalSDK}.

\section{Methodology}
\label{sec_method}

In this project, we aim to perform sentiment analysis using three different modalities—textual, acoustic, and visual extracted from video content. This is done using a unified architecture that uses pre-trained BERT models, specifically ‘bert-base-uncased’ to encode each modality individually and then apply an early fusion mechanism for sentiment prediction. The following subsections discuss it in detail.

\subsection{Modality-specific Feature Representation}

\subsubsection{Text Modality}

The textual data, i.e., transcriptions of the spoken content are tokenized using the BERT tokenizer. These tokens are then passed into the pre-trained BERT-base model (uncased). We extract the hidden state of the special [CLS] token from the final layer to obtain a sentence-level embedding. This representation encapsulates the contextual and semantic information of the entire utterance, capturing nuances like sentiment-bearing words, negations, and discourse structures.

\subsubsection{Acoustic Modality}

As discussed, audio features are extracted using the COVAREP toolkit \cite{degottex2014covarep}, which computes a comprehensive set of low-level descriptors, including Mel-frequency cepstral coefficients (MFCCs), pitch, glottal source parameters \cite{drugman2012detection,alku2002normalized,alku1997parabolic}, and voicing-related features. These frame-level features are arranged in a temporal sequence and treated as a pseudo-token sequence. We employ a BERT-based encoder adapted for continuous audio inputs to process this sequence. The final embedding is derived by mean pooling over the sequence.

\subsubsection{Visual Modality}

The processed visual data stream is passed to a BERT encoder to yield a high-dimensional vector representation of the visual sentiment behavior. As with the acoustic data, we use mean pooling over the final embedding to get the input sequence of visual data.

\subsection{Early Fusion}

Once the data streams are processed, we merge them using a technique called early fusion. In early fusion, the features from each modality are combined before they are passed to the classifier \cite{snoek2005early}. In this project, once we extract fixed-length embeddings from each modality using pre-trained BERT-based encoders, we concatenate these vectors to form a single joint feature vector:

\begin{equation}
    H_{fused} = [ H_{text} ; H_{audio} ; H_{visual} ]
\end{equation}

This allows the model to jointly learn the interdependencies across modalities. For instance, it can capture phenomena where the textual content appears positive, but the tone or facial expression indicates sarcasm or negativity. Early fusion helps the model contextualize such cross-modal interactions more effectively than late fusion, where predictions are made independently and merged afterward \cite{snoek2005early}. The features are merged using cross-attention layers.

Moreover, for future experimentation, we plan to employ late fusion, where predictions from different models at a later stage are combined. This means that each modality is processed independently and the final decision is made by aggregating the outputs of the independent models. We will then compare the results of early fusion and late fusion to determine which strategy is more effective. But, for this project, we stick to the early fusion mechanism.

\subsection{Classification}

After early fusion, the concatenated feature vector is passed through a classification head responsible for transforming the representation into final sentiment prediction. The architecture of this classification component includes the following stages:

\subsubsection{Fully Connected Layers}
A linear transformation is applied to the fused vector, followed by a ReLU activation. This enables the model to capture higher-order interactions among the fused features and learn non-linear representations necessary for accurate classification.

\subsubsection{Layer Normalization}
Applied after the dense transformation to stabilize the learning process and normalize feature distributions across the batch. This helps accelerate convergence and improves generalization.

\subsubsection{Dropout Layer}
A dropout mechanism (with p = 0.3) is employed post-activation to reduce overfitting by randomly zeroing a portion of neurons during training, encouraging robustness in the learned features.

\subsubsection{Softmax Output Layer}
The final dense layer maps the hidden representation to a vector of length 7, corresponding to the 7 sentiment classes (ranging from –3 to +3). A softmax activation is applied to convert these logits into a probability distribution, and the class with the highest probability is selected as the model's prediction.

\subsection{Training Configurations}

To train and assess the model effectively for sentiment classification, we employed cross-entropy loss. The Adam optimizer is used due to its efficiency and adaptive learning rate capabilities. A learning rate of 1e-4 ensures stable convergence without overshooting. A mini-batch size of 32 was selected to balance computational efficiency with training stability, especially considering the resource demands of BERT models. A dropout rate of 0.3 is applied after the dense layer and within the encoders to improve generalization. Training monitors the validation accuracy and loss and stops when performance plateaus. The early stopping patience was set to 10 epochs. This avoids overfitting and saves computational resources by selecting the best-performing model checkpoint. The model was trained for 50 epochs, however, the training stopped at the 25th epoch due to an early stopping mechanism with the best validation loss as 0.0182. For this model, we had 8 transformer layers and 16 attention heads. Also, the random seed was set to 42 for reproducibility.

\section{Results}
\label{sec_res}

To assess the effectiveness of our proposed multimodal sentiment analysis model, we evaluated its performance on the training, validation, and test sets using a comprehensive set of metrics. These include Mean Absolute Error (MAE), 7-class Accuracy, 7-class F1 Score, Binary Accuracy, and Binary F1 Score. Additionally, the training dynamics were monitored over 25 epochs to ensure stable convergence and generalization.

\subsection{Training}

As shown in Figure 2, the training loss exhibited a steady decline over the course of training, indicating consistent learning. Meanwhile, the validation loss remained low and stable throughout, showing no signs of overfitting. This behavior is further corroborated by the strong performance on unseen validation and test samples, as described below.

\begin{figure*}[!t] 
    \centering
    \includegraphics[width=\textwidth]{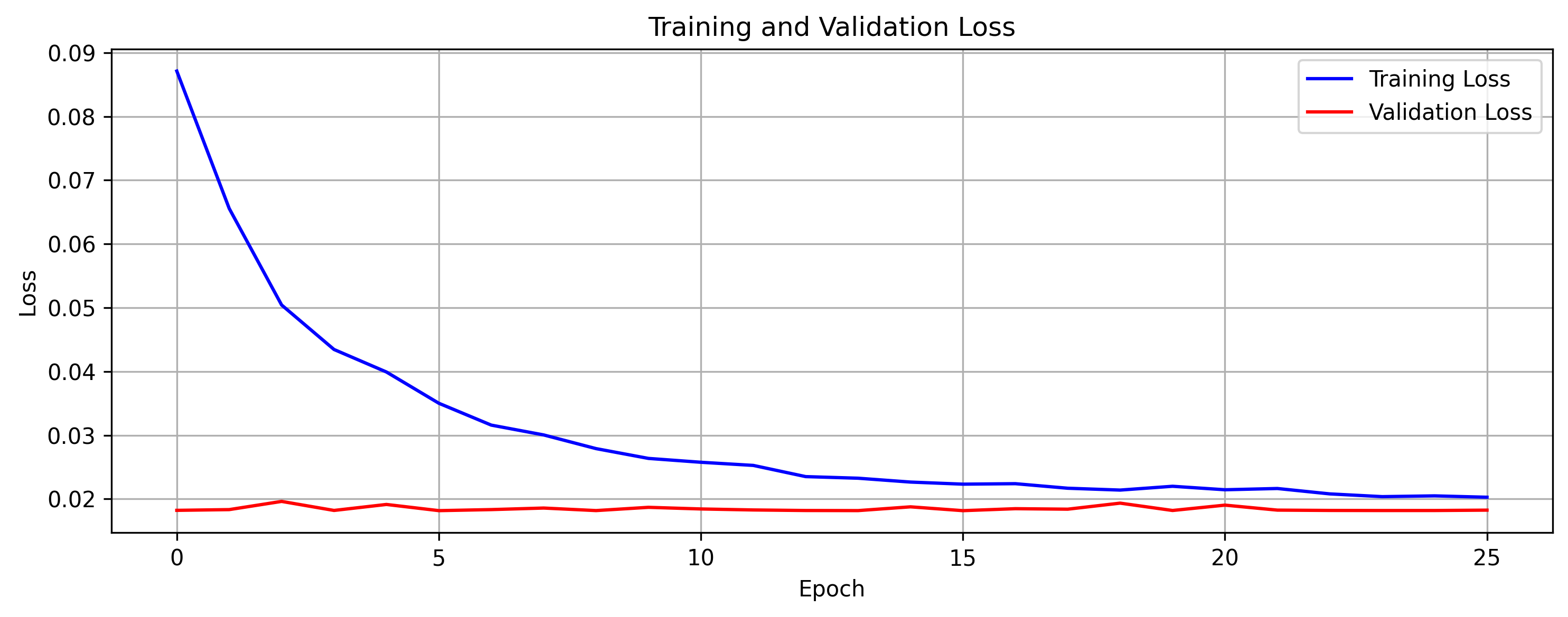}
    \caption{Training and validation loss curves over 25 epochs.}
    \label{fig:train_val_loss_curve}
\end{figure*}

At epoch 16, the model achieved its best performance on the validation set and was saved as the final model checkpoint. The train metrics at that point are described in table 2:

\begin{table}[h!]
    \centering
    \begin{tabular}{|p{0.7\linewidth}|l|}
        \hline
        \textbf{Metric} & \textbf{Value} \\
        \hline
        Mean Absolute Error (MAE) & 0.1115 \\
        7-class Accuracy & 96.79\% \\
        7-class F1 Score & 0.9522 \\
        Binary Accuracy & 94.94\% \\
        Binary F1 Score & 0.9740 \\
        \hline
    \end{tabular}
    \caption{Train metrics at epoch 16 (best model).}
    \label{tab:train_res}
\end{table}

Moreover, the corresponding loss values at that point are shown in table 3:

\begin{table}[h!]
    \centering
    \begin{tabular}{|p{0.7\linewidth}|l|}
        \hline
        \textbf{Loss Type} & \textbf{Value} \\
        \hline
        Training Loss & 0.0223 \\
        Validation Loss & 0.0182 \\
        \hline
    \end{tabular}
    \caption{Training and validation loss values at epoch 16  (best model).}
    \label{tab:train_val_loss}
\end{table}

\subsection{Validation}

The model exhibited strong performance on the validation set, indicating its ability to generalize beyond the training data. At the best epoch (Epoch 16), the metrics were:

\begin{table}[h!]
    \centering
    \begin{tabular}{|p{0.7\linewidth}|l|}
        \hline
        \textbf{Metric} & \textbf{Value} \\
        \hline
        Mean Absolute Error (MAE) & 0.1080 \\
        7-class Accuracy & 97.35\% \\
        7-class F1 Score & 0.9604 \\
        Binary Accuracy & 96.59\% \\
        Binary F1 Score & 0.9827 \\
        \hline
    \end{tabular}
    \caption{Validation metrics at epoch 16 (best model).}
    \label{tab:val_res}
\end{table}

These metrics reflect a high degree of fidelity in both fine-grained sentiment classification (across 7 classes from –3 to +3) and binary sentiment polarity detection (positive vs. negative).

\subsection{Test}

Upon evaluation on the unseen test set using the best model checkpoint, the model maintained its high performance across all metrics:

\begin{table}[h!]
    \centering
    \begin{tabular}{|p{0.7\linewidth}|l|}
        \hline
        \textbf{Metric} & \textbf{Value} \\
        \hline
        Mean Absolute Error (MAE) & 0.1060 \\
        7-class Accuracy & 97.87\% \\
        7-class F1 Score & 0.9682 \\
        Binary Accuracy & 95.44\% \\
        Binary F1 Score & 0.9767 \\
        \hline
    \end{tabular}
    \caption{Test set performance of the best model checkpoint.}
    \label{tab:test_res}
\end{table}

The consistently high accuracy and F1 scores on both the validation and test sets demonstrate the model’s robustness and reliability.

\section{Conclusion}
\label{sec_concl}

In this project, we presented a transformer-based multimodal sentiment analysis model that uses an early fusion of textual, acoustic, and visual features to effectively predict sentiment on the CMU-MOSEI dataset. By combining modality-specific transformer-based encoders with a classification head enhanced by Layer Normalization, Dropout, and ReLU activation, the model achieved strong generalization and high accuracy across both binary and 7-class sentiment prediction tasks.

Our approach demonstrated consistent improvements across validation and test sets, achieving 97.87\% accuracy and 0.9682 F1-score on the 7-class sentiment test set, alongside a low MAE of 0.1060. These results highlight the model’s ability to not only learn cross-modal relationships effectively but also avoid overfitting through regularization techniques and efficient model design.

The model's performance shows the importance of early fusion strategies when combined with pre-trained BERT encoders and proper regularization for multimodal tasks. Future work may explore the integration of attention-based fusion mechanisms or a comparison between early fusion and late fusion techniques to further enhance interpretability and performance.

\end{document}